# Using WordNet for Building WordNets

Xavier Farreres, German Rigau, Horacio Rodríguez
Departament de Llenguatges i Sistemes Informàtics.
Universitat Politècnica de Catalunya. Barcelona. Spain.
{farreres, horacio, g.rigau}@lsi.upc.es

**Abstract**

This paper summarises a set of methodologies and techniques for the fast construction of multilingual WordNets. The English WordNet is used in this approach as a backbone for Catalan and Spanish WordNets and as a lexical knowledge resource for several subtasks.

## 1 Motivation and Introduction

One of the main issues in last years as regards NLP activities is the increasingly fast development of generic language resources. A lot of such resources, including both software and lingware items (lexicons, lexical databases, grammars, corpora marked in several ways) have been made available for research and industrial applications.

Special interest presents, for knowledge-based NLP tasks, the availability of wide coverage ontologies. Most known ontologies (as GUM, CYC, ONTOS, MICROKOSMOS, EDR or WORDNET, see [Gomez 98] for an extensive survey) differ in great extent on several characteristics (e.g. broad coverage vs. domain specific, lexically oriented vs. conceptually oriented, granularity, kind of information placed in nodes, kind of relations, way of building, etc.). It is clear, however, that for a wide range of applications, WordNet (WN) [Miller 90] as become a de-facto standard.

The success of WordNet has determined the emergence of several projects that aim the construction of WordNets for other languages than English (e.g., [Hamp & Feldweg 97], [Artale et al. 97]) or to develop multilingual WordNets (the most important project in this line is EuroWordNet (EWN)[1]).

The construction of a WN for a language Lg (LgWN) can be tackled in different ways according to the lexical sources available. Of course the manual construction can be undertaken quite straightforwardly and leads to the best results in terms of accuracy, but has the important drawback of its cost. So, other approaches have been carried out taking profit of available resources in fully automatic or semi-automatic ways.

Which are these lexical resources? Basically four kinds of resources have been used: 1) English WN (EnWN), as an initial skeleton for trying to attach the words of Lg to it, 2) already existing taxonomies of Lg (both at word and at sense level), 3) bilingual (English and Lg) and 4) monolingual (Lg) dictionaries. All the approaches using EnWN as skeleton are based on the assumption of a close conceptual similarity between English and Lg, in such a way that most of the structure (relations) in EnWN could be maintained for LgWN.

In the case of bilingual dictionaries the usual approach is to try to link the English counterpart of entries to synsets in EnWN and to assume that the entry can be linked to the same synset.

Monolingual dictionaries have been used basically as a source for extracting taxonomic (hypernym) links between words (or senses [Bruce & Guthrie 92], [Rigau et al. 97]) and in lower extent for extracting other kinds of semantic relations [Richardson 97] (e.g. meronymic links).

Once a taxonomy of Lg (already existing or built from a monolingual MRD) is available, the task can consist of 1) enriching the taxonomic structure with other semantic links (manually or automatically), as is the case of building individual WNs, or 2) merging this structure with other already existing ontologies (as EnWN or EWN).

This paper presents our approach to the construction of WNs for two languages, Spanish and Catalan, and linking the first one to EWN. We have developed a methodology that uses as core source EnWN[2]. The methodology implies 1)

---

[1] http://www.let.uva.nl/~ewn/ The aim of EWN project is to build a multilingual database with WordNets for several european languages (in the first phase, Dutch, Italian and Spanish in addition to English).

[2] We have used WordNet 1.5.

The use of EnWN for guiding the selection of the basic concepts of our WNs, 2) the use of EnWN as skeleton for linking Spanish and Catalan words to English synsets using bilingual dictionaries, 3) the use of EnWN, together with bilingual and monolingual dictionaries for allowing the construction of taxonomies (at sense level) of our languages and 4) the use of EnWN together with already built fragments of SpWN and CtWN for merging and incorporating these taxonomies to our WNs.

In section 2 an overall description of our approach is given. Sections 3, 4 and 5 focus on the procedures for extracting connections between words/senses/synsets. Section 3 is devoted to procedures based on the use of bilinguals, section 4 on the construction of taxonomies and section 5 deals with the merging method. In all these sections we will enphasize the role played by EnWN as Knowledge Source. Section 6, finally, presents some conclusions of our work.

## 2 Our way of building WordNets

As we have pointed out in the introduction, our aim has been to design a methodology (and a software environment supporting it) for facilitating the task of building WNs from our sources. As we are involved in EWN project (covering the Spanish part), the methodology has been defined to be compatible which the general approach, guidelines and landmarks of the whole project but also to allow a parallel development of the CtWN.

The general approach for building EWN is described in [Vossen et al. 97]. Roughly speaking, the approach follows a top-down strategy trying to assure a high level of overlapping between languages, at least in the highest levels of the hierarchy, but reflecting the language-specific lexicalizations and providing the maximum of freedom and flexibility for building the individual WordNets. Basically it consists of three major steps: 1) Construction of core-WordNets for a set of common base concepts (around 800 nouns and 200 verbs), 2) enrichment of these sets providing relational links and incorporating their direct semantic contexts and 3) top-down extension of these core-WordNets.

In our case two different approaches have been followed for dealing with nouns and verbs[3].

In the case of verbs most of the work has been performed manually. The main source of information has been the Pirapides database [Castellón et al. 97] that consists of 3,600 English verbs forms organized around Levin's Semantic Classes connected to WN1.5 senses. The database contains the theta-Grids specifications for each verb (its semantic structure in terms of cases or thematic roles), translation to Spanish and Catalan forms[4] and diathesis information. The connections extracted from this database were cross-validated with the information provided by bilingual dictionaries in order to improve their accuracy.

In the case of nouns we have followed EWN strategy in the next way:

1) The two highest levels of EnWN (top concepts and direct hyponyms) were manually translated into Spanish (including variants). The results were filtered dropping out words appearing less than five times as genus terms in our monolingual dictionary [DGILE 87] or occurring less than 50 times in DGILE definition corpus[5] and less than 100 times in LEXESP corpus[6].

This initial set (Spanish core concepts, 361 synsets) was then compared with base concept sets of other sites of EWN (roughly the union of intersection pairs between languages was considered as the common base concepts set). The missing concepts in Spanish were manually added and vertically bottom up extended leading to the common Base Concept set (around 800 synsets). Catalan Base Concepts set was then built to cover the Spanish Base Concepts set.

2) The enrichment of the BC set has been performed in two steps. First, using bilinguals as main lexical source, and then using other sources (mainly taxonomies). These processes are described below.

## 3 Using English WordNet with bilinguals

When trying to build a lexical taxonomy from scratch, we can take profit of a preexisting lexical taxonomy, EnWN in our case, assuming it is well formed, as a skeleton of a taxonomy where we will fill in the lexical data. This ensures several advantages: it speeds up the construction of a large lexicon, as the only problem left is the decision where to attach the lexical data. There are also some problems: nobody ensures that the wellformedness of a lexical taxonomy for a language keeps true for another language, there must be semantic closeness between both languages. We have therefore assumed that the structure of the WN taxonomy would suffice in the earlier stages of the construction of the our WNs.

---

[3] Although other categories can be included in EWN (and cross-category relations an be established) only nouns and verbs have been introduced until now in our WordNets except for demostration purposes.

[4] Spanish and Catalan are languages close enough for allowing a simultaneous development of lexical sources.
[5] i.e. set of all definitions included in DGILE (1 million words)
[6] balanced corpus of Spanish (5 Million words).

So, we need to choose synonyms in Spanish[7] for the English words present in the original synsets of WN. One way to fulfil our requirements is using bilingual dictionaries (see [Knight & Luk 94], [Okumura & Hovy 94]). But we have to perform a sense disambiguation task in order to know which sense of both words (the Spanish and the English one) is being referred. In other words, we have to decide, for which sense of the Spanish word and for which synset in WordNet a relation of synonymy is being defined.

There is also another minor problem to overcome, the unification of the two directions of the bilingual dictionary, which in few cases are symmetrical, to collect all translations together. It is true that unifying both directions of the bilingual dictionary implies loss of information potentially important (e.g. the order in which translations are written is relevant). But the lack of systematic work in the construction of the bilinguals makes this information of very doubtful utility.

Thus, we have processed the bilinguals creating what we have called the homogeneous bilingual, which is a bilingual with both directions mixed. Then, for each Spanish word, we have collected all the words given as correct translations. And this has been the source for our work of attachment of Spanish words to WordNet synsets.

Having collected all the translations of a Spanish word together, we have then classified the words in classes depending on their behaviour. They can be classified in three dimensions: polysemy, structural and conceptual.

In the **polysemy** dimension, we classify the words in classes depending on the number and kind of translations. For example, all entries that have only one translation fall in the same class when this translation is monosemous in WN terms; all entries that have several translations fall in another class when these translations are polysemous.

---

[7]Although we ilustrate the methodology considering only Spanish, we performed the whole process for both Catalan and Spanish (and we provide results for both).

In the **structural** dimension, we classify the words in classes depending on the relation that the translations owns in WN. For example, all entries which have several translations, sharing some of them a common synset in WN, fall in the same category; all entries in which one translation is a direct hyponym of other translation fall in the same category, etc.

In the **conceptual** dimension, we apply the conceptual distance formula (which is explained in section 4.2.1.) on elements of the entries. For example, all entries with a low conceptual distance between synsets of their translations fall in the same class.

Each of these classes defines a set of entries with the same behaviour. A confidence score has been assigned to each class by means of a manual validation of a significant sample extracted from them. We decided to accept the classes with a precission of 85% or more as classes of words to include in the first version of SpWN.

Bilinguals can be used a step further stating a supposition: when several methods give the same result for the same Spanish word, the confidence for this attachment increases. We have carried out an experiment checking the classes in pairs, evaluating the precission of the set of intersections, and in all cases the precission increased. We have removed the cases where the precision was over 85%, the threshold applied in the previous experiment. This caused an increment of 40% of the original set of attachments.

Furthermore, it is clear that if we merge more bilinguals, the homogeneous resulting will be larger, and will then generate larger classes. But, what is even more important, the classes are more precise because some bilinguals lack the inclusion of some translations for some words. Table 1 shows the current figures of both CtWN and SpWN following this approach (see [Atserias et al. 97] and [Benítez et al. 98] for further details of the whole process and tools used).

| Nouns | Words | Synsets | Connections |
|---|---|---|---|
| **Spanish** | 23,217 | 18,578 | 41,293 |
| **Catalan** | 5,231 | 4,723 | 7,193 |
| **Verbs** | | | |
| **Spanish** | 3,087 | 3,219 | 7,960 |
| **Catalan** | 3,337 | 3,219 | 9,078 |

Table 1: current volumes of CtWN and SpWN.

The last point to address is the extension of the intersection method to larger number of classes. If with two classes the intersection increased the confidence an equivalent increase when intersecting larger numbers of classes can be expected.

As a matter of fact, the extension of the intersection method would be nothing more than performing a multivariant statistical analysis, where each of the classes would be a factor. The interesting result of this multivariant analysis would be a formula which could be used to calculate the value of the confidence of an attachment, depending on the number of classes in which it occurs.

## 4 Building Taxonomies using WordNet

### 4.1 Exploiting taxonomies from MRDs

A straightforward way of obtaining a LgWN can be performed acquiring taxonomic relations from conventional dictionaries following a purely bottom up strategy. That is, 1) parsing each definition for obtaining the genus, 2) performing a genus disambiguation procedure, and 3) building a natural classification of the concepts as a concept taxonomy with several tops. Following this purely descriptive methodology, the semantic primitives of the LgWN could be obtained by collecting those dictionary senses appearing at the top of the complete taxonomies derived from the dictionary. By characterizing each of these tops, the complete LgWN could be produced. For DGILE, the complete noun taxonomy was derived using the automatic method described by [Rigau et al. 97][8].

However, several problems arise due to a) the source (i.e., circularity, errors, inconsistencies, omitted genus, etc.) and b) the limitation of the genus sense disambiguation techniques applied (i.e., [Bruce et al. 92] report 80% accuracy using automatic techniques, while [Rigau et al. 97] report 83%). Furthermore, the top dictionary senses do not usually represent the semantic subsets that the LgWN needs to characterize in order to represent useful knowledge for NLP systems. In other words, there is a mismatch between the knowledge directly derived from an MRD and the knowledge needed by a LgWN.

To illustrate the problem we are facing, let us suppose we plan to place the FOOD concepts in the LgWN. Neither collecting the taxonomies derived from a top dictionary sense (or selecting a subset of the top dictionary senses of DGILE) closest to FOOD concepts (e.g., substancia -substance-), nor collecting those subtaxonomies starting from closely related senses (e.g., bebida -drinkable liquids- and alimento -food-) we are able to collect exactly the FOOD concepts present in the MRD. The first are too general (they would cover non-FOOD concepts) and the second are too specific (they would not cover all FOOD dictionary senses because FOODs are described in many ways).

All these problems can be solved using a mixed methodology. That is, by attaching selected top concepts (and its derived taxonomies) to prescribed semantic primitives represented in the LgWN. Thus, first, we prescribe a minimal ontology (represented by the semantic primitives of the LgWN) able to represent the whole lexicon derived from the MRD, and second, following a descriptive approach, we collect, for every semantic primitive placed in the LgWN, its subtaxonomies. Finally, those subtaxonomies selected for a semantic primitive are attached to the corresponding LgWN semantic category.

We used as semantic primitives the 24 lexicographer's files (or semantic files) into which the 60,557 noun synsets (87,641 nouns) of WN are classified[9]. Thus, we considered the 24 semantic tags of WN as the main LgWN semantic primitives to which all dictionary senses must be attached. In order to overcome the language gap we also used a bilingual Spanish/English dictionary.

### 4.2 Attaching DGILE dictionary senses to semantic primitives

In order to classify all nominal DGILE senses with respect to WordNet semantic files, we used a similar approach to that suggested by [Yarowsky 92]. This task is divided into three fully automatic consecutive subtasks. First, we tag a subset (due to the difference in size between the monolingual and the bilingual dictionaries) of DGILE dictionary senses by means of a process that uses the conceptual distance formula (see 4.2.1); second, we collect salient words for each semantic file; and third, we enrich each DGILE dictionary sense with a semantic tag collecting evidence from the salient words previously computed.

---

[8]This taxonomy contains 111,624 dictionary senses and has only 832 dictionary senses which are tops of the taxonomy (these top dictionary senses have no hypernyms), and 89,458 leaves (which have no hyponyms). That is, 21,334 definitions are placed between the top nodes and the leaves.

[9]One could use other semantic classifications, such as Roget's Thesaurus [Yarowsky 92], the LDOCE semantic or pracmatic codes [Slator 91] or even better, a Spanish semantic classification such as the "Diccionario Ideológico de la Lengua Española J. Casares" (DILEC). Really, when using this methodology a minimal set of informed seeds are needed. These seeds can be collected from MRDs, thesauri or even by introspection. (see [Yarowsky 95]).

### 4.2.1 Attaching WordNet synsets to DGILE headwords.

For each DGILE definition, the conceptual distance between headword and genus has been computed using WN1.5 as a semantic net. We obtained results only for those definitions having English translations (using a bilingual dictionary) for both headword and genus. By computing the conceptual distance between two words (w1,w2) we are also selecting those concepts (c1i,c2j) which represent them and seem to be closer with respect to the semantic net used. Conceptual distance is computed using formula (1).

$$(1) \quad dist(w_1, w_2) = \min_{\substack{c_{1i} \in w_1 \\ c_{2j} \in w_2}} \sum_{c_k \in path(c_{1i}, c_{2j})} \frac{1}{depth(c_k)}$$

That is, the conceptual distance between two concepts depends on the length of the shortest path[10] that connects them and the specificity of the concepts in the path.

In this way, we obtained a preliminary version of 29,205[11] dictionary definitions semantically labelled (that is, with WN lexicographer's files) with an accuracy of 64% (61% at a sense level). That is, a corpus (collection of dictionary senses) classified in 24 partitions (each one corresponding to a semantic category).

### 4.2.2 Collecting the salient words for every semantic primitive.

Thus, we can collect the salient words (that is, those representative words for a particular category) using a Mutual Information-like formula (2), where w means word and SC semantic class.

$$(2) \quad AR(w, SC) = \Pr(w|SC) \log_2 \frac{\Pr(w|SC)}{\Pr(w)}$$

Intuitively, a salient word[12] appears significantly more often in the context of a semantic category than at other points in the whole corpus, and hence is a better than average indicator for that semantic category. The words selected are those most relevant to the semantic category, where relevance is defined as the product of salience and local frequency. That is to say, important words should be distinctive and frequent.

We performed the training process considering only the content word forms from dictionary definitions[13] and we discarded those salient words with a negative score. Thus, we derived a lexicon of 23,418 salient words (one word can be a salient word for many semantic categories).

### 4.2.3 Enriching DGILE definitions with WordNet semantic primitives.

Using the salient words per category (or semantic class) gathered in the previous step we labelled the DGILE dictionary definitions again.

When any of the salient words appears in a definition, there is evidence that the word belongs to the category indicated. If several of these words appear, the evidence grows. We add together their weights, over all words in the definition, and determine the category for which the sum is greatest, using formula (3).

$$(3) \quad W(SC) = \sum_{w \in definition} AR(w, SC)$$

Thus, we obtained a second semantically labelled version of DGILE. This version has 86,759 labelled definitions (covering more than 93% of all noun definitions) with an accuracy rate of 80% (we have gained, since the previous labelled version, 62% coverage and 16% accuracy).

Although we used the 24 lexicographer's files of WordNet as semantic primitives, a more fine-grained classification could be made. For example, all FOOD synsets are classified under **<food, nutrient>** synset in file 13. However, FOOD concepts are themselves classified into 11 subclasses (i.e., **<yolk>**, **<gastronomy>**, **<comestible, edible, eatable, ...>**, etc.). Thus, if the LgWN we are planning to build needs to represent **<beverage, drink, potable>** separately from the concepts **<comestible, edible, eatable, ...>** a finer set of semantic primitives should be chosen, for instance, considering each direct hyponym of a synset belonging to a semantic file also as a new semantic primitive or even selecting

---

[10]We only consider hypo/hypermym relations.
[11]Due to the different sizes of the dictionaries used we only compute the conceptual distance for 31% of the noun dictionary senses.
[12]Instead of word lemmas, this study has been carried out using word forms because word forms rather than lemmas are representative of typical usages of the sublanguage used in dictionaries.

[13]After discarding functional words.

for each semantic file the level of abstraction we need.

### 4.3 Selecting the main top beginners for a semantic primitive

This section is devoted to the location of the main top dictionary senses for a given semantic primitive in order to correctly attach all its subtaxonomies to the correct semantic primitive in the LgWN.

In order to illustrate this process we will locate the main top beginners for the FOOD dictionary senses. However, we must consider that many of these top beginners are structured. That is, some of them belong to taxonomies derived from other ones, and then cannot be directly placed within the FOOD type. This is the case of vino (wine), which is a zumo (juice). Both are top beginners for FOOD and one is a hyponym of the other.

First, we collect all genus terms from the whole set of DGILE dictionary senses labelled in the previous section with the FOOD tag (2,614 senses), producing a lexicon of 958 different genus terms (only 309, 32%, appear more than once in the FOOD subset of dictionary senses).

As the automatic dictionary sense labelling is not free of errors (around 80% accuracy)[14] we can discard some senses by using filtering criteria.

- Filter 1 (F1) removes all FOOD genus terms not assigned to the FOOD semantic file during the mapping process between the bilingual dictionary and WN.
- Filter 2 (F2) selects only those genus terms which appear more times as genus terms in the FOOD category. That is, those genus terms which appear more frequently in dictionary definitions belonging to other semantic tags are discarded.
- Filter 3 (F3) discards those genus terms which appear with a low frequency as genus terms in the FOOD semantic category. That is, infrequent genus terms (given a certain threshold) are removed. Thus, F3>1 means that the filtering criteria have discarded those genus terms appearing in the FOOD subset of dictionary definitions less than twice.

At the same level of genus frequency, filter 2 (removing genus terms which are more frequent in other semantic categories) is more accurate than filter 1 (removing all genus terms the translation of which cannot be FOOD). For instance, no error appears when selecting those genus terms which appear 10 or more times (F3) and are more frequent in that category than in any other (F2), discarding only 3% of correct genus terms (see [Rigau et al. 98] for complete figures).

### 4.4 Automatically building large scale taxonomies from DGILE

The automatic Genus Sense Disambiguation task in DGILE has been performed following [Rigau et al. 97]. This method reports 83% accuracy when selecting the correct hypernym by combining eight different heuristics using several methods and types of knowledge (two of the heuristics use WN).

Once the main top beginners (relevant genus terms) of a semantic category are selected and every dictionary definition has been disambiguated, we collect all those pairs labelled with the semantic category we are working on having one of the genus terms selected. Using these pairs we finally build up the complete taxonomy for a given semantic primitive. That is, in order to build the complete taxonomy for a semantic primitive we fit the lower senses using the second labelled lexicon and the genus selected from this labelled lexicon.

Although, both final taxonomic structures produce more flat taxonomies than if the task is done manually, a few arrangements could be done at the top level of the automatic taxonomies. Studying the main top beginners we can easily discover an internal structure between them (for FOOD, 18 or 48 depending on the criteria selected).

Performing the process for the whole dictionary we obtained for F2+(F3>9) a taxonomic structure of 35,099 definitions and for F2+(F3>4) the size grows to 40,754. Testing the results on FOOD taxonomies we achived 99% accuracy with the first criterion and 96% with the second.

## 5 Extending and Filling Gaps.

Up to now we have described a methodology to connect words from a language to a WN skeleton, and another methodology to build taxonomies.

The words finally connected in the first process, apart from the precission threshold criterion, do not follow any other criterion: they are not the most important, neither the topmost nor the lowermost concepts in the hierarchy; the connections are scattered all over the skeleton. The final set of words connected to the skeleton is random, and we don't have any control over it.

---

[14] Most of them are not really errors. For instance, all fishes must be ANIMALs, but some of them are edible (that is, FOODs). Nevertheless, all fishes labelled as FOOD have been considered mistakes.

Furthermore, we also find relevant words not connected to the hierarchy.

We are currently developing a methodology which tries to fill the gaps by merging the taxonomy automatically extracted, and the sparse skeleton. By now we have studied very simple and short structures.

We have then two hierarchies to compare, and two ways of connecting them: the already extracted connections (**A**) between Spanish words and synsets, and the translations (**B**) given by the bilinguals (not disambiguated). We have then looked for the next simple configurations:

(4) sp - en
    |   |
    sp - en

where Spanish words are connected between them via the automatically extracted taxonomy, and the English words via WN. The English words can be connected to the Spanish via A or via B, or they can be unconnected. Then we obtain eight configurations. We have evaluated up to now three of these classes:

- class 1: connections via A above and below[15]
- class 2: connections via A above and B below
- class 4: connections via B above and A below

Below there is a table showing volumes. The experiment was carried out on four file senses which in our opinion would differ in their behaviour: food and artifact, which are classified very similarly in Spanish and in English, and mental process and communication, which are not so clear:

| semfile | class 1 | class 2 | class 4 |
|---|---|---|---|
| artifact | 222 | 560 | 224 |
| mental process | 54 | 182 | 105 |
| communiccation | 119 | 270 | 198 |
| food | 30 | 66 | 56 |

Table 2: class volumes.

Of these volumes, some were already extracted with the previous methods, but some are newly produced connections. Some of the already existing connections were incorrect, and led to incorrect deductions. Of the newly added connections, a sample has been evaluated, giving the results below:

---

[15]This class simply provides additional evidence over the confidence score.

| semfile | class 1 | class 2 | class 4 |
|---|---|---|---|
| artifact | 99% | 77% | 89% |
| mental process | 99% | 77% | 79% |
| communication | 99% | - | 78% |
| food | 99% | - | 68% |

Table 3: overall results

| semfile | class 2 | class 4 |
|---|---|---|
| artifact | 50% | 85% |
| mental process | 50% | 65% |
| communication | - | 50% |
| food | - | 74% |

Table 4: results for new connections

We can then decide, after studying all the cases, to accept the connections above a threshold, or we can also try to combine them to extract more precise ones. For example, some promising combinations could be:

(5) sp - en -- A
    |   |
    sp - en -- B
    |   |
    sp - en -- A

(6) sp - en -- A
    |   |
    sp -en --no connection
    |   |
    sp - en -- A

which are the combinations of classes 2 and 4 in (5), and combinations of two new classes in (6). Furthermore, we are planning to apply an iterative bootstrapping method taking profit of those links with higher confidence scores gathered in previous steps (acting as anchors) to spread evidence where no connections have been found.

We are also considering the possibility of, not only filling gaps in the middle levels of the hierarchy, but also to extend the LgWN adding subtaxonomies to bottom synsets of WN, trying to cover semantic fields specific of Lg not covered by the original WN.

## 6. Conclusions

An approach for building in a fast and automatic way substantial fragments of WordNets have been presented. The method uses as skeleton English WordNet and extracts its knowledge from a variety of lexical sources (taxonomies, monolingual and bilingual dictionaries). Our approach makes extensive uses of English WordNet in several steps of the building process. The system has been applied to build Spanish WordNet (within the framework of EuroWordNet) and Catalan WordNet. First, following [Atserias et al. 97], we applied a set of complementary techniques for linking Spanish and Catalan words collected from a bilingual MRDs (for nouns) and lexicons (for verbs) to English WordNet. Second, by applying the methodology described in section 4 we are able to build up accurate taxonomies from monolingual MRDs (see [Rigau et al. 97] and [Rigau et al. 98]). Third, taking profit of both lexical resources (the sparse connections produced by the first methodology and the taxonomies produced by the second) we have presented a novel bootstrapping methodology for covering substantial parts of the new WordNets not covered previously.

## Acknowledgements

This research has been partially funded by the Spanish Research Department (ITEM Project TIC96-1243-C03-03), the Catalan Research Department (CREL project), and the UE Comision (EuroWordNet LE4003).